\documentclass{article}
\usepackage{iclr2027_conference,times}
\usepackage[hyphens]{url}
\usepackage{graphicx}
\usepackage{amsmath,amssymb}
\usepackage{booktabs,array,tabularx}
\usepackage{xcolor}
\usepackage{caption}
\newcolumntype{Y}{>{\raggedright\arraybackslash}X}
\usepackage[hidelinks]{hyperref}
\hypersetup{pdftitle={Beyond Outcome Rewards: Constructing and Assigning Retrieval Credit for Search Agents},pdfauthor={},pdfsubject={Retrieval reward construction and action credit assignment}}

\newcommand{\ind}{\mathbf{1}}
\iclrfinalcopy
\usepackage{etoolbox}
\makeatletter
\patchcmd{\@maketitle}{Published as a conference paper at ICLR 2027}{Preprint}{}{\PackageError{arxiv}{Could not replace conference header}{}}
\makeatother
\title{Beyond Outcome Rewards: Constructing and\\Assigning Retrieval Credit for Search Agents}
\author{%
Wenyu Huang\textsuperscript{\dag},
Xinyu Hou\textsuperscript{\ddag},
Pavlos Vougiouklis\textsuperscript{\ddag},
Ruofei Lai\textsuperscript{\ddag},
Jeff Z. Pan\textsuperscript{\dag}\\
\textsuperscript{\dag}University of Edinburgh, UK\quad
\textsuperscript{\ddag}Huawei Technologies Research \& Development (UK) Limited\\
\texttt{w.huang@ed.ac.uk}}
\hypersetup{pdfauthor={Wenyu Huang, Xinyu Hou, Pavlos Vougiouklis, Ruofei Lai, Jeff Z. Pan}}

\begin{document}
\maketitle
\begin{abstract}
Search agents enable Large Language Models (LLMs) to iteratively retrieve and use information for complex multi-hop questions. Reinforcement Learning with Verifiable Rewards (RLVR) offers a promising approach for post-training such agents, but its reliance on sparse, outcome-based supervision can make credit assignment difficult and limit learning efficiency. In this paper, we systematically investigate how intermediate supervision can improve reinforcement learning for search agents. We study a range of reward-shaping and credit-assignment strategies that provide learning signals from intermediate retrieval steps. Building on these insights, we develop a training framework that combines intermediate signals with final outcome rewards to improve learning from multi-step search trajectories. Experiments across multiple benchmarks under matched training conditions demonstrate improvements in aggregate search-agent performance and show that both the choice of intermediate signal and where its credit is assigned affect training behaviour. These findings show that reward design and credit assignment are important design dimensions for training effective search agents.
\end{abstract}
\section{Introduction}
Retrieval-augmented generation (RAG) \citep{DBLP:conf/nips/LewisPPPKGKLYR020} allows large language models to answer questions using external evidence. For complex questions, the required evidence may be spread across several documents, and an initial retrieval may reveal what the model needs to search for next. Search agents such as Search-o1 \citep{li2025searcho1} address this problem by interleaving reasoning with retrieval, using each observation to guide subsequent actions. Search-R1 \citep{jin2025searchr} and ReSearch \citep{chen2025researchlearningreasonsearch} use reinforcement learning with verifiable rewards (RLVR) to train this behaviour from answer feedback. Through interaction with a search environment, the policy learns which queries to issue and how to use the returned information to answer the question.

A final outcome reward, however, provides limited information about the searches along the way. Two trajectories can both receive zero answer reward even though one retrieves part of the required evidence and the other retrieves none. When all rollouts in a question group receive the same answer score, group-relative training supplies no outcome advantage, leaving differences in retrieval progress unused. Intermediate supervision can make this progress available to the learner before it produces a correct final answer. Yet a retrieval score alone does not specify how to update the policy: the same evidence gain can change the reward for the whole trajectory or provide credit to the particular search that produced it.

We therefore investigate how intermediate retrieval supervision depends on its credit assignment. We study two connected choices: what retrieved information should earn a reward, and how that reward should enter the policy update. A search can retrieve a fact before finding the evidence needed to establish its prerequisites. Rewarding that fact immediately recognises evidence already acquired, while waiting for its prerequisites rewards progress through the annotated dependency structure. These choices can assign credit to different searches even when the agent eventually retrieves the same evidence. Adding that credit to the final reward changes the shared advantage for the whole trajectory, whereas assigning it locally changes the update on a particular search action. Comparing retrieval signals under both formulations lets us examine whether a signal's benefit persists across these choices.

To support this study, we develop a data construction pipeline that composes corpus-grounded relations into questions while retaining intermediate answers, supporting passages and evidence dependencies. The intermediate answers and supporting passages identify which subquestions a retrieval helps answer, while the dependency structure records how those subquestions connect. Retaining both lets us change the rule for crediting evidence while holding the questions and their grounding fixed. The same annotations also support analysis of what a trained agent retrieves, separately from whether its final answer is correct.

\begin{figure}[t]
\centering
\includegraphics[width=\linewidth]{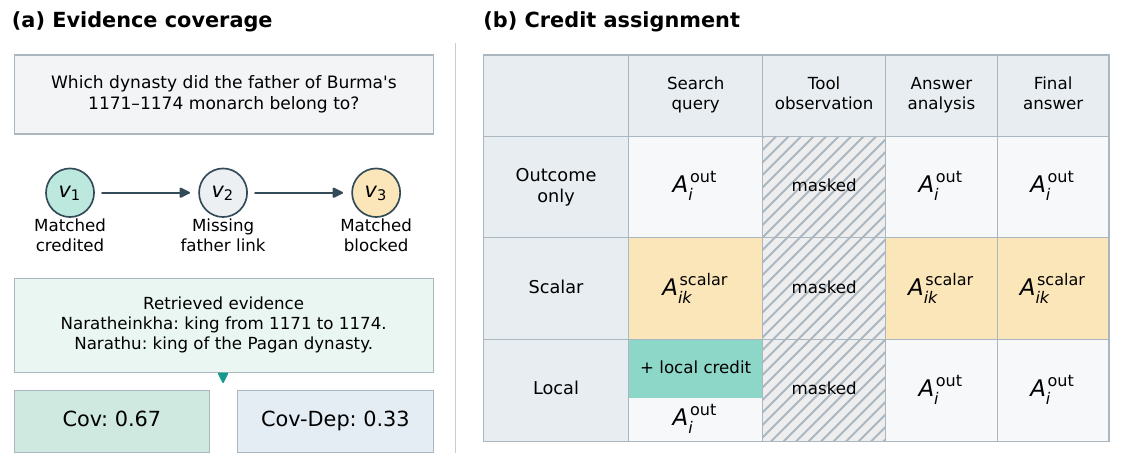}
\caption{Observed coverage and schematic credit assignment. (a) An observed search matches $v_1$ and $v_3$; the missing father link at $v_2$ blocks dependency credit for $v_3$, giving Cov 0.67 and Cov-Dep 0.33. Text is abridged. (b) Scalar changes the shared advantage; local retains the outcome advantage and adds credit on executed query tokens (Equations~\ref{eq:scalar}--\ref{eq:local}). Tool observations are masked. Colour shows support, not magnitude.}
\label{fig:pipeline}
\end{figure}

Using these annotations, we compare three retrieval signals on the same observations. Grounded evidence coverage (Cov) measures how much of the annotated evidence has been retrieved; coverage with dependencies (Cov-Dep) requires the same evidence to satisfy prerequisite relations; answer match (AM) checks for a reference answer in the retrieved text. Each signal supplies either a scalar reward bonus or a local, group-relative advantage alongside the final outcome reward (Figure~\ref{fig:pipeline}). Within local credit, we further test whether feedback needs to remain aligned with its source search and whether its benefit is concentrated in groups with tied outcome rewards.

We evaluate these choices with Qwen3-4B across seven QA benchmarks. Both scalar and local retrieval supervision improve on outcome-only training, with local credit giving the largest gains. Cov-local improves average F1 by 3.09, with a larger gain of 4.81 on multi-hop QA. Our controlled comparisons further show that the effectiveness of a retrieval signal depends on how its credit is incorporated and whether it remains aligned with the action that produced it. These results show that retrieval rewards should be designed together with the way their credit enters the policy update.

We further compare Cov-local with outcome-only training on annotated development questions. Cov-local retrieves more supporting evidence, while the two policies show similar coverage-conditioned answer F1. This suggests that the main change lies in evidence acquisition.

\section{Constructing corpus-grounded retrieval rewards}
\label{sec:construction}
Given a question $x$, a search agent produces a trajectory $\tau=(a_1,o_1,\ldots,a_T,o_T,y)$. Each search action $a_t$ uses the question and previous observations to retrieve passages $o_t$, which guide subsequent searches and the final answer $y$.

We construct questions with annotated subquestions and supporting passages to measure the evidence acquired along a trajectory. Below, we describe the data and define three retrieval signals and their per-search increments.

\subsection{Training data and evidence annotations}
\label{sec:data}
To evaluate intermediate retrieval during training, we construct questions together with their supporting evidence. A final answer alone cannot identify which intermediate facts a search has recovered. Our construction therefore preserves the subquestions, their dependencies and their source passages alongside each question, providing annotations for retrieval supervision as well as outcome evaluation. Following MuSiQue \citep{trivedi-etal-2022-musique} in composing questions with dependent reasoning steps, we begin with entity pairs that co-occur within a KILT sentence and are linked by a Wikidata relation \citep{DBLP:journals/cacm/VrandecicK14}. We retain the source passage for each relation, allowing the evidence to be traced back to Wikipedia as relations are combined into questions.

We compose these relations into chains and conjunctive structures. For chains, we favour relation types that usually map a subject to one object, and require each selected subject--property pair to have exactly one object. This object can then serve as the subject of a subsequent step. For conjunctive structures, each branch must admit multiple candidates while their intersection identifies a single answer. Within the selected relations, neither branch alone determines the answer, so their combination adds a constraint rather than repeating an already sufficient relation. The resulting decomposition forms a graph in which each node represents a subquestion with its own answer, and directed edges record dependencies on earlier answers. A node can therefore require evidence from multiple passages. For example, ``Which university did both person A and person B attend?'' combines two education relations into one subquestion. We attach both supporting passages to this node, whereas a node representing a single relation retains that relation's source passage.

Once the structure is selected, GPT-5.6 Luna generates a descriptive reference for each root entity without naming it. Templates use these references and the selected relations to assemble the question, reference answer aliases, subquestions and dependency edges. The graph determines the answer and decomposition; the generated description supplies a way to refer to the starting entity in the question. For each subquestion, we also record the locations of its supporting passages and annotate mentions of its answer within them. These locations connect the generated question to the retriever's chunks: a node is covered when retrieved chunks match all of its required passages. The graph edges specify which earlier answers a subquestion depends on, while the attached passages specify the evidence needed to answer it. Both coverage signals below use these annotations; answer match uses only the original question's reference answer aliases.

We then exclude records with missing grounding, detected graph shortcuts, or unmatched structural categories, and sample deterministically from ten categories covering single-evidence, conjunctive, chain, and fan-in questions. This produces 14,000 training questions shared by all conditions. The training mixture contains 2,500 one-hop, 3,300 two-hop, 4,900 three-hop and 3,300 four-hop questions, grouped by their construction templates. The one-hop subset includes 1,200 single-evidence and 1,300 conjunctive questions; the remaining questions comprise 5,400 chain, 1,600 wide and 4,500 fan-in structures. The mixture therefore includes both sequential dependencies and evidence combined across branches. These categories describe annotated structures, not the number of searches an agent must execute.

Retaining both passage-level evidence and dependencies lets us score the same retrieved observations with or without prerequisite constraints, while holding the questions and grounding fixed. We also reserve a separate set of annotated questions for the evidence-coverage analysis in Section~\ref{sec:dev_coverage}. The construction therefore supports both reward comparisons and analysis of search behaviour.\footnote{Appendix~\ref{app:data} provides construction constraints, mixture counts, and details of passage matching and grounding checks.}

\subsection{Retrieval signals}
\paragraph{Grounded evidence coverage (Cov).}
Let $V_x$ be the subquestion nodes for question $x$ and $D_{\leq t}$ the documents observed through search action $t$. Following Section~\ref{sec:data}, let $c_v(D_{\leq t})$ equal one when the retrieved chunks match all required supporting passages of node $v$, and zero otherwise. Cov measures the fraction of covered nodes:
\begin{equation}
 \phi_{\mathrm{Cov}}(t)=|V_x|^{-1}\sum_{v\in V_x}c_v(D_{\leq t}).
 \label{eq:unordered}
\end{equation}
Cov credits a covered node regardless of whether evidence for its predecessor nodes has been retrieved.

\paragraph{Coverage with dependencies (Cov-Dep).}
Cov-Dep uses the same passage-matching rule but credits a covered node only after its prerequisite nodes have also been credited. Let $\mathrm{pa}(v)$ denote these prerequisites; root nodes have $\mathrm{pa}(v)=\varnothing$. Starting from an empty credited set, repeatedly add covered nodes whose prerequisites are already in the set. The resulting set $S_t$ is the least fixed point
\begin{equation}
 S_t=\{v\in V_x:c_v(D_{\leq t})=1,\ \mathrm{pa}(v)\subseteq S_t\},
 \qquad \phi_{\mathrm{Cov\text{-}Dep}}(t)=\frac{|S_t|}{|V_x|}.
 \label{eq:graph}
\end{equation}
Both coverage signals require all passages of a conjunctive node and give each node equal weight. Their difference is whether dependencies between nodes constrain credit. For a chain $A\!\to\!B\!\to\!C$, retrieving evidence for $A$ and $C$ gives Cov $2/3$ but Cov-Dep $1/3$. Retrieving $B$ later unlocks both $B$ and the already covered $C$. Thus closure can change the timing and magnitude of credit without requiring the agent to retrieve evidence in graph order.

\paragraph{Answer match (AM).}
AM checks whether a retrieved document contains a reference answer alias for the original question. It compares normalised strings in the document title and body. The potential $\phi_{\mathrm{AM}}(t)$ becomes one after the first match and remains zero until then. AM needs no subquestion graph or evidence-location annotations, but an answer string can occur in an unrelated passage. All three signals use completed tool observations.

\paragraph{Retrieval-progress events.}
The potentials above measure cumulative progress. To identify what each search adds, we convert them into nonnegative increments over the previous maximum. For $X\in\{\mathrm{Cov},\mathrm{Cov\text{-}Dep},\mathrm{AM}\}$:
\begin{equation}
 e^X_t=\max\!\left(0,\phi_X(t)-\max_{0\leq u<t}\phi_X(u)\right),\qquad \phi_X(0)=0.
 \label{eq:event}
\end{equation}
Repeated retrieval of already credited evidence produces a zero retrieval-progress event. We use this event definition for all three signals. Their final accumulated progress can be equal even when credit arrives at different searches. Section~\ref{sec:credit} describes how these events enter the policy update.

\section{Assigning retrieval credit to policy updates}
\label{sec:credit}
Retrieval events indicate which searches acquire new evidence, but using these events for learning also requires deciding where to assign credit. We compare two approaches: scalar incorporation combines retrieval progress with the final outcome reward for the whole trajectory, while local incorporation adds retrieval credit to the search action that produced the evidence.

Both approaches build on Group Relative Policy Optimisation (GRPO) \citep{grpo}. For each question, the policy samples a group of trajectories, and their outcome rewards are standardised within the group to obtain relative advantages. Let $r_i$ be the outcome reward for trajectory $i$, $k$ index response tokens, and $z(\cdot)$ denote this group standardisation. Under outcome-only GRPO, the advantage $A_i^{\mathrm{out}}=z(r)_i$ is shared by all trainable response tokens in trajectory $i$. We retain the same clipped policy objective and change the advantages supplied to it: scalar incorporation changes this shared advantage, while local incorporation retains it and adds an action-specific term (Figure~\ref{fig:pipeline}b).

\paragraph{Scalar incorporation.}
Writing $e^X_{it}$ for the retrieval event at action $t$ in trajectory $i$, scalar incorporation uses
\begin{equation}
 R_i^X=r_i+\alpha\sum_t e^X_{it},\qquad
 A_{ik}^{\mathrm{scalar}}=z(R^X)_i.
 \label{eq:scalar}
\end{equation}
Here $\alpha$ scales the accumulated retrieval bonus, which equals the final maximum potential before scaling. Ordinary GRPO then broadcasts the resulting advantage to trainable response tokens, with no local residual. Retrieval differences can change trajectory rankings and can supply variance even when the original answer scores are tied.

\paragraph{Action-aligned local incorporation.}
Local incorporation keeps the scalar outcome score $r_i$. At each search step $t$, we centre retrieval events across eligible rollouts for the same question. Local credit requires valid action-to-token alignment and at least two eligible rollouts with nonzero variance in their retrieval events. We standardise each retrieval signal and then the combined local component. With one active signal, these two passes give $\ell^X_{it}=z(z(e^X_{\cdot t}))_i$ with both standardisation passes using the same eligible set. Ineligible or constant groups have zero local residual.

Let $\mathcal{E}_i$ contain the eligible retrieval-action indices in trajectory $i$. For $t\in\mathcal{E}_i$, let $Q_{it}$ be the nonempty set of trainable executed-query tokens and $L_{it}$ the trainable length of the corresponding full tool-call payload. We use the full tool-call length to set the event mass $M_{it}=L_{it}\ell^X_{it}$, then distribute this mass across $Q_{it}$. The resulting token advantage is
\begin{equation}
 A^{\mathrm{local}}_{ik}=A_i^{\mathrm{out}}+
 \lambda\sum_{t\in\mathcal{E}_i}\ind[k\in Q_{it}]\frac{L_{it}\ell^X_{it}}{|Q_{it}|}.
 \label{eq:local}
\end{equation}
The coefficient $\lambda$ controls the strength of local retrieval credit. This credit is signed: a below-group-average event can receive negative credit. The outcome advantage remains on all trainable response tokens, including those receiving local credit. When no retrieval event is eligible, the local sum is zero. Tool responses are excluded from the policy loss. The residual can reinforce or oppose the outcome advantage. Equation~\ref{eq:local} preserves the total event mass when distributing it across tokens; the full tool-call length sets its scale.

\paragraph{Controlled comparisons.}
We further conduct controlled comparisons within the local formulation to examine event--action correspondence and which outcome groups benefit from retrieval credit. First, the permutation conditions test whether retrieval credit needs to remain aligned with the action that produced it. They shuffle nonzero signed event masses across rollouts for the same question at the same search step, preserving the available credit while breaking event--action correspondence. Second, the outcome-group restrictions test whether local credit is useful mainly when the outcome reward cannot distinguish trajectories. For both Cov and AM, flat-only retains local credit only in groups with tied answer scores, where the outcome advantage is zero; no-flat retains it only in groups with differing scores. Both retain the outcome advantage and leave the remaining local updates unscaled. These restrictions therefore compare where local credit is applied while also reducing its total amount.\footnote{Appendix~\ref{app:hyper} gives signal-matching rules, eligibility and normalisation checks, and details of permutation and outcome-group gating.}

\section{Experimental design and results}
\label{sec:experiments}
\paragraph{Training and evaluation.}
We train Qwen3-4B-Instruct-2507 \citep{qwen3technicalreport} on the same 14,000 questions. The searchable corpus is the KILT-2019 Wikipedia corpus \citep{petroni2021kilt}, partitioned into over 29 million 100-word chunks. Retrieval uses E5-base-v2 \citep{wang2022text} and returns three chunks per query. The retriever remains fixed during policy optimisation. Training uses GRPO with five rollouts per question and the same training budget across conditions. The outcome reward is answer F1, $r_i=\mathrm{F1}(y_i,y^*)$. Our baseline, outcome-only training (OO), uses this reward without retrieval supervision. All conditions share the clipped policy objective and KL regularisation, differing only in the reward or token advantage supplied to the objective. Group standardisation uses the sample standard deviation with $\epsilon=10^{-6}$. All results use the final checkpoint and greedy decoding. Appendix~\ref{app:hyper} gives the remaining settings.

We use a common scalar coefficient $\alpha=0.25$ to add retrieval supervision while giving outcome reward greater weight in the combined score. Offline checks show that this setting preserves most outcome rankings on saved trajectories (Appendix~\ref{app:calibration}). For local incorporation, we use a fixed coefficient $\lambda=0.5$ across all conditions, keeping the scaling coefficient constant when comparing signals and credit-assignment rules.

We evaluate on 2WikiMultiHopQA \citep{ho-etal-2020-constructing}, HotpotQA \citep{yang2018hotpotqa}, MuSiQue \citep{trivedi-etal-2022-musique}, Bamboogle \citep{press-etal-2023-measuring}, Natural Questions \citep{kwiatkowski-etal-2019-natural}, TriviaQA \citep{joshi-etal-2017-triviaqa}, and PopQA \citep{mallen-etal-2023-trust}. We use the sampled evaluation sets provided by \citet{zheng2025deepresearcherscalingdeepresearch}. Each dataset contains 512 evaluation examples except Bamboogle, which contains 125. We primarily report answer F1 in the main text and provide exact match (EM) results in Appendix~\ref{app:results}. Avg denotes the micro average over all 3,197 examples, giving each question equal weight. This benchmark suite informed development decisions; we use the separate 1,200-question synthetic development split to analyse evidence coverage (Section~\ref{sec:dev_coverage}).

\begin{table}[t]
\caption{Answer F1 (\%) on all seven datasets. Avg summarises performance across all seven datasets. Cov measures grounded evidence coverage; Cov-Dep adds dependency constraints to the same coverage signal. AM uses retrieved-answer matching. Bold marks the highest value in each column. The blocks compare scalar and local incorporation for each signal. Figure~\ref{fig:mechanisms} reports the local-credit interventions; complete F1 and EM results for all 13 conditions appear in Appendix~\ref{app:results}.}
\label{tab:main}
\centering\small\setlength{\tabcolsep}{3pt}
\begin{tabular}{lrrrrrrrr}
\toprule
Method & 2Wiki & HotpotQA & MuSiQue & Bamboogle & NQ & TriviaQA & PopQA & Avg \\
\midrule
OO & 57.30 & 53.74 & 24.93 & 57.73 & 42.04 & 74.46 & 53.45 & 51.25 \\
\midrule
Cov-scalar & 59.26 & 57.54 & 25.69 & 59.51 & 43.08 & 74.09 & 54.50 & 52.64 \\
Cov-Dep-scalar & 57.81 & 58.42 & 25.11 & 59.54 & \textbf{44.51} & 74.58 & \textbf{54.88} & 52.82 \\
AM-scalar & 59.27 & 57.11 & 25.89 & 60.21 & 43.25 & 74.45 & 54.16 & 52.66 \\
\midrule
Cov-local & \textbf{62.82} & 58.91 & \textbf{28.84} & \textbf{61.91} & 43.43 & 75.86 & 54.37 & \textbf{54.34} \\
Cov-Dep-local & 60.96 & \textbf{59.49} & 25.49 & 58.29 & 42.01 & 76.09 & 52.42 & 52.96 \\
AM-local & 59.26 & 59.13 & 27.23 & 59.01 & 43.23 & \textbf{76.31} & 54.57 & 53.51 \\
\bottomrule
\end{tabular}
\end{table}

\subsection{The benefit of a signal depends on its incorporation}
\label{sec:signal_interaction}
Table~\ref{tab:main} shows that local incorporation outperforms scalar incorporation for each retrieval signal, with the largest gain for evidence coverage. Cov-local achieves 54.34 average F1, compared with 51.25 for outcome-only training. All three scalar variants also improve over OO, but assigning credit locally yields further gains whose size depends on the retrieval signal. The signal ranking also changes: Cov is strongest under local incorporation, whereas Cov-Dep is strongest under scalar incorporation.

Comparing Cov with Cov-Dep shows how dependency constraints affect scalar and local training differently. Both use the same evidence nodes and grounding, but Cov credits a newly covered node before its prerequisites are retrieved. Removing dependency closure raises local F1 from 52.96 to 54.34, whereas scalar F1 changes from 52.82 to 52.64 (Figure~\ref{fig:mechanisms}a). The two formulations use retrieval events differently: scalar incorporation sums them before group normalisation, while local incorporation compares them at each search step. Even when two event sequences end at the same coverage, they can produce different local updates. This distinction helps explain why signal construction must be considered together with incorporation.

\subsection{Performance across tasks}
\begin{figure}[t]
\centering\includegraphics[width=\linewidth]{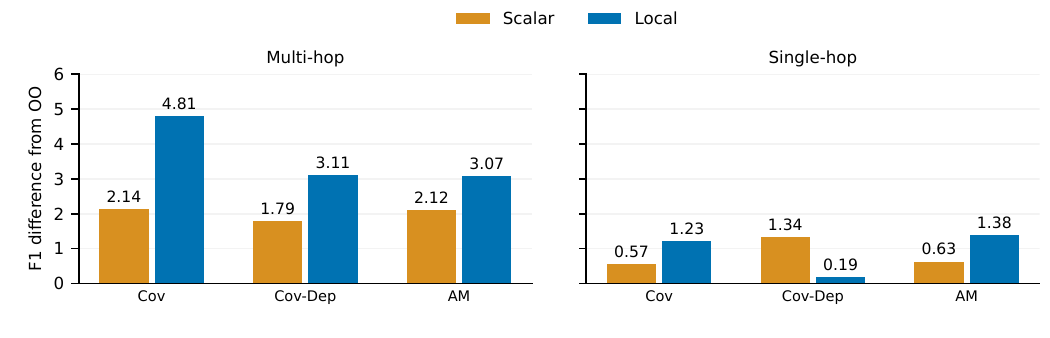}
\caption{Scalar and local credit relative to OO across task groups. Each score averages over the group's evaluation questions. Appendix~\ref{app:taskgroups} gives full scores and benchmark composition.}
\label{fig:taskgroups}
\end{figure}

Figure~\ref{fig:taskgroups} shows that both Cov-local and AM-local improve multi-hop and single-hop QA, with larger gains on multi-hop QA. Cov-local has a larger advantage on multi-hop QA, while AM-local slightly exceeds it on single-hop QA. The clearer distinction is on multi-hop QA, where tracking evidence for intermediate subquestions yields a larger gain than matching the final answer string. The benefit of local incorporation nevertheless depends on the signal and task group. Cov-Dep-local improves over its scalar counterpart on multi-hop QA but falls below it on single-hop QA. Both Cov-local and AM-local improve over OO on all seven benchmarks, extending their benefits across both task groups.

\section{Which properties of retrieval credit matter?}
\label{sec:analysis}
We examine how preserving event--action correspondence and choosing which outcome groups receive retrieval credit affect performance. We then analyse evidence acquisition by the trained policies on the annotated development questions.
\begin{figure}[t]
\centering\includegraphics[width=\linewidth]{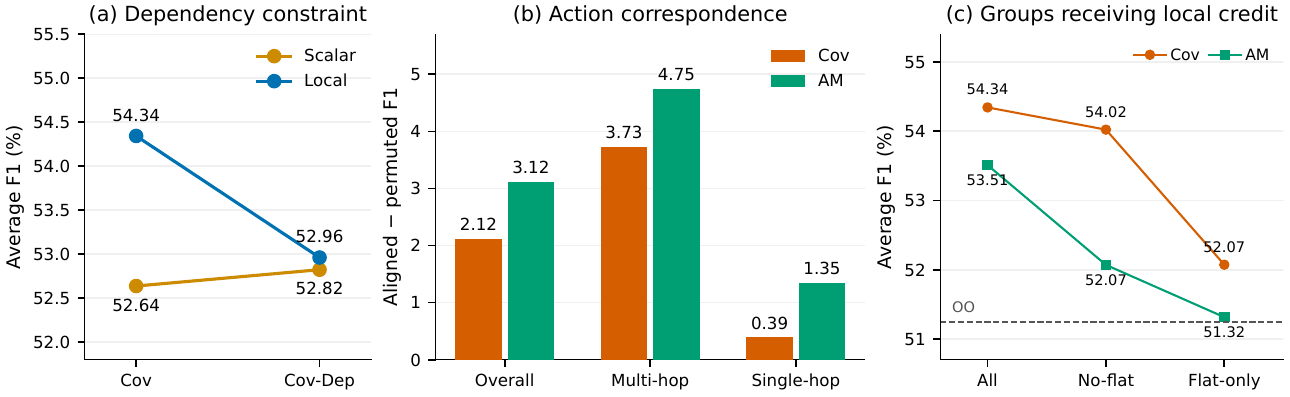}
\caption{F1 comparisons under three retrieval-credit interventions. (a) Removing closure helps local incorporation, but not scalar incorporation. (b) Aligned minus permuted F1 overall and within each task group; positive values favour aligned credit. (c) Using both outcome groups gives the highest aggregate F1 for Cov and AM; the dashed line denotes OO.}
\label{fig:mechanisms}
\end{figure}

\subsection{Preserving action correspondence improves local credit}
\label{sec:correspondence}
Local credit links a search action to the evidence it retrieves. We test whether this link matters by comparing aligned credit with permuted credit for both Cov and AM. In the aligned condition, each eligible search receives the credit computed from its own retrieval event. In the permuted condition, we shuffle nonzero signed event masses across rollouts for the same question at the same search step. Credit still applies to query tokens, and the shuffle preserves the signed event masses; it changes which search receives each credit.

Figure~\ref{fig:mechanisms}b shows that aligned credit outperforms permuted credit for both signals, increasing average F1 by 2.12 for Cov and 3.12 for AM. The advantage is larger on multi-hop QA than on single-hop QA. This pattern suggests that associating feedback with the search that produced it is particularly useful when answering a question requires several pieces of evidence.

Compared with OO, Cov-permute still achieves higher F1 (52.22 versus 51.25), while AM-permute performs worse (50.39). Coverage supervision therefore retains some benefit even when credit is reassigned, but preserving the correspondence raises F1 further to 54.34. For both signals, assigning credit to its source search is more effective than randomly reassigning it among searches.

\subsection{Restricting local credit to outcome ties limits its benefit}
\label{sec:flat_groups}
For both Cov and AM, we compare local credit applied to all rollout groups with flat-only, which retains it only in groups with tied answer scores, and no-flat, which retains it only in groups with differing scores.

For both signals, using all groups gives the highest aggregate F1, followed by no-flat and then flat-only (Figure~\ref{fig:mechanisms}c). For Cov, excluding tied-outcome groups reduces F1 by 0.32, whereas retaining only those groups reduces it by 2.27. AM follows the same ordering, with losses of 1.44 and 2.19, respectively. Thus useful retrieval supervision is not confined to groups where the outcome reward cannot distinguish trajectories. Cov-flat-only remains above OO, while AM-flat-only performs similarly to OO. Both restrictions reduce the amount of local credit as well as changing which groups receive it.

\subsection{Evidence acquisition}
\label{sec:dev_coverage}
\begin{figure}[t]
\begin{minipage}[t]{0.49\linewidth}
\centering\includegraphics[width=\linewidth]{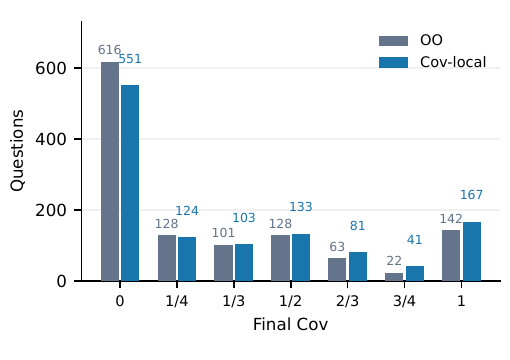}
\caption{Final coverage distribution for OO and Cov-local on the same 1,200 development questions. Bars give the number of questions at each Cov value.}
\label{fig:dev_coverage}
\end{minipage}\hfill
\begin{minipage}[t]{0.49\linewidth}
\centering\includegraphics[width=\linewidth]{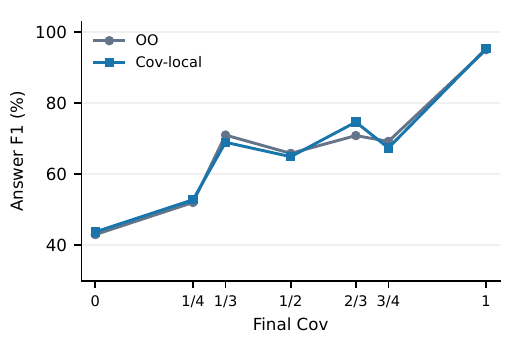}
\caption{Mean answer F1 at each final Cov value. Each policy is grouped separately, so the curves can contain different question IDs at the same coverage.}
\label{fig:dev_coverage_f1}
\end{minipage}
\end{figure}

We compare OO and Cov-local on the same 1,200 synthetic development questions using the benchmark evaluation settings. The grounded annotations let us measure how much of the required evidence each policy retrieves. Figure~\ref{fig:dev_coverage_f1} also plots answer F1 conditional on coverage for each policy. Appendix~\ref{app:dev_coverage} describes the trajectory analysis.

Across both policies, higher evidence coverage is associated with higher answer F1 (Figure~\ref{fig:dev_coverage_f1}), linking the retrieval signal to answer quality. Figure~\ref{fig:dev_coverage} shows that local coverage supervision shifts more questions towards higher evidence coverage. Cov-local reduces the number of zero-coverage questions from 616 to 551 and increases complete coverage from 142 to 167 questions. More questions therefore acquire annotated evidence, including the full set of required passages, although complete coverage remains uncommon.

Mean Cov increases from 27.51\% under OO to 31.97\% under Cov-local. The policies average 2.45 and 2.55 searches per question, respectively.

The two Cov--F1 curves follow a similar overall pattern: at the same coverage level, Cov-local does not consistently achieve higher F1. Together with the coverage distribution, this suggests that query-level supervision primarily improves evidence acquisition, with less indication of improved use of the retrieved evidence. The groups at each coverage level differ in both size and question identity, so the curves do not isolate evidence use on the same questions. Further work could examine whether supervision beyond search queries improves how agents use retrieved evidence.

\section{Related work}
\paragraph{Multi-turn search agents.}
RAG \citep{DBLP:conf/nips/LewisPPPKGKLYR020} combines language generation with evidence from an external corpus. Multi-turn agents extend this interaction by deciding what to retrieve as reasoning proceeds. ReAct \citep{yao2023react} provides a general framework for interleaving reasoning and actions. Search-o1 \citep{li2025searcho1} integrates search into the reasoning process to retrieve missing knowledge. Search-R1 \citep{jin2025searchr} and ReSearch \citep{chen2025researchlearningreasonsearch} train search behaviour with reinforcement learning and answer feedback. An empirical study by \citet{jin2025empiricalstudyreinforcementlearning} reports limited gains from intermediate retrieval rewards in the settings examined. We study how their effectiveness depends on the way retrieval credit enters the policy update.

\paragraph{Intermediate supervision.}
\citet{lightman2023let} study process supervision for mathematical reasoning, providing feedback on intermediate steps rather than only the final result. For retrieval, R3-RAG \citep{li2024r3rag} combines answer correctness with document-relevance rewards. SWiRL \citep{goldie2025swirl} decomposes multi-step reasoning and tool-use trajectories into action-level sub-trajectories for data filtering and RL optimisation. More recent search-agent methods assess different aspects of retrieval progress. HiPRAG \citep{wu2026hiprag} evaluates search necessity and adds a hierarchical process bonus to outcome and format rewards. BiCAA \citep{huang2026bicaa} combines forward answer-solvability gains with hindsight estimates of a search step's importance for the final answer. Our study uses corpus-grounded annotations to compare answer matching with evidence coverage. The two coverage variants share their grounding and differ in whether prerequisite relations constrain credit, allowing us to trace that choice through the resulting updates.

\paragraph{Credit assignment for retrieved evidence.}
An intermediate signal also requires a rule for assigning credit to the policy. STAMP \citep{xu2026stamp} verifies cited evidence against a reference graph, attributes it to the action that first exposed it, and modulates the outcome advantage without changing its sign. GDCR \citep{liu2026gdcrsapo} rewards newly retrieved and cited entities according to their graph distance to the answer; SAPO standardises these rewards across steps within each trajectory and scales their contribution by the absolute outcome advantage. CW-GRPO \citep{wang2026cwgrpo} uses an LLM judge to assess retrieval utility and reasoning correctness, redistributing outcome advantage across successful search rounds while assigning uniform weights to failed trajectories.

These methods scale intermediate credit by the outcome advantage. We compare retrieval events across rollouts of the same question at the same search step and add the resulting signed residual to selected tokens. When eligible events vary, this residual can remain active at zero outcome advantage and can reverse a nonzero advantage. Our experiments examine whether its source action and activation on tied-outcome groups affect performance. The scalar comparison also shows that learning from outcome ties is shared by both forms of intermediate supervision (Appendix~\ref{app:calibration}).

\section{Limitations}
Our analysis examines evidence acquisition and retrieval credit. A further question is how retrieved evidence enters subsequent reasoning. This requires tracing evidence use through intact search trajectories, including cases where additional coverage does not change the answer. Such an analysis could inform supervision that connects evidence acquisition with evidence use.

The interaction between reward construction and update strength also warrants closer study. Experiments that vary event timing, reward magnitude and credit eligibility independently could distinguish their contributions, while comparisons under matched update strength could clarify the role of normalisation in scalar and local incorporation. Extending these controls across model sizes and retrievers would test whether these credit-assignment behaviours generalise.

\section{Conclusion}
Our study shows that intermediate retrieval supervision should be designed together with its credit assignment. Using corpus-grounded questions with intermediate evidence annotations, we find that the same retrieval signal has different effects depending on how it enters the policy update. Dependency constraints help distinguish this interaction: requiring prerequisites benefits scalar incorporation, while crediting evidence as it is acquired works better under local incorporation. A more constrained evidence signal therefore does not necessarily provide more useful supervision.

Across our experiments, local retrieval credit is most effective when it remains associated with the search that produced the evidence and complements outcome supervision across both tied and differing outcome groups. Its value extends beyond supplying an update when answer rewards are tied. On the development questions, higher coverage and similar coverage-conditioned F1 suggest that local supervision primarily changes evidence acquisition, leaving evidence use as a separate question. Together, these findings support treating evidence scoring and action-level credit assignment as coupled design choices when training search agents.

\label{main-text-end}
\clearpage
\section*{Reproducibility statement}
Section~\ref{sec:data} describes corpus-grounded question generation; Sections~\ref{sec:construction} and~\ref{sec:credit} define the retrieval rewards and credit assignment. Appendix~\ref{app:hyper} provides training settings and reward-scale checks; Appendix~\ref{app:data} describes data construction and grounding checks. Appendix~\ref{app:results} provides complete final-checkpoint EM and F1 results and dataset sensitivity; Appendix~\ref{app:coverage_example} details the example in Figure~\ref{fig:pipeline}a. We will release the code, trained models and constructed dataset upon acceptance.

\section*{AI use statement}
In this work, we used generative AI tools for synthetic data generation and to assist in writing experimental code. GPT-5.6 Luna generated descriptive root references for the synthetic questions, while graph operations and templates assembled the question structures, answers and evidence dependencies. Additionally, we used generative AI tools to assist manuscript drafting and editing. We have reviewed all AI-assisted work, including the generated code and manuscript text. Experimental results were checked against saved evaluation outputs. We take responsibility for the final content of this work, including text, claims or artifacts produced with the aid of generative AI.

\bibliography{references}
\bibliographystyle{iclr2027_conference}
\clearpage
\appendix
\section{Experimental details}
\label{app:hyper}
\subsection{Hardware, optimisation and runtime}
We train Qwen3-4B-Instruct-2507 on a single node with two NVIDIA A100 GPUs, each with 80\,GB memory. The implementation uses VERL \citep{DBLP:conf/eurosys/ShengZYWZZPL025} for distributed optimisation and SGLang \citep{zheng2024sglang} for rollout generation, with tensor parallelism 1 and bfloat16 rollout generation. The fixed E5-base-v2 retriever described in Section~\ref{sec:experiments} uses a FAISS IndexFlatIP index. The training hyperparameters below are shared across conditions.

\begin{table}[!htbp]
\caption{Shared training hyperparameters.}
\centering\small
\begin{tabularx}{\linewidth}{lY}
\toprule
Setting & Value \\
\midrule
Optimiser & AdamW \\
Learning rate & $10^{-6}$ \\
Learning-rate schedule & Warmup followed by a constant learning rate \\
Warmup ratio & 0.285 \\
Rollouts per question & 5 \\
Training batch size & 128 \\
PPO epochs per rollout batch & 1 \\
Policy-ratio clipping interval & $[0.8,1.2]$ \\
Dual-clip coefficient & 3.0 \\
KL loss coefficient & 0.001 \\
Training sampling temperature & 1.0 \\
\bottomrule
\end{tabularx}
\end{table}

The KL penalty is applied in the policy loss rather than added to the reward.

\paragraph{Runtime.}
Across seven measured conditions, training takes approximately 28--31 hours. These times include rollout generation, policy updates, periodic validation and checkpoint saving, but exclude startup and the initial validation.

\subsection{Prompt and tool interface}
\label{app:prompt}
We use the same system instruction and user template for training, the seven-benchmark evaluation and the synthetic development evaluation. The question replaces \texttt{\{question\}} below. We reproduce the text verbatim, with line wrapping for readability. The system message includes the tool instructions and function schema supplied by the model's chat template.

\paragraph{System message (including tool template)}
\begin{quote}\small\ttfamily\raggedright
You are a helpful and harmless assistant.\par
\smallskip
\# Tools\par
\smallskip
You may call one or more functions to assist with the user query.\par
\smallskip
You are provided with function signatures within \textless{}tools\textgreater{}\textless{}/tools\textgreater{} XML tags:\par
\textless{}tools\textgreater{}\par
\{"type": "function", "function": \{"name": "search", "description": "Searches the web for relevant information based on the given query.", "parameters": \{"type": "object", "properties": \{"query\_list": \{"type": "array", "description": "A list of fully-formed semantic queries. The tool will return search results for each query.", "enum": null, "items": null, "minItems": null, "maxItems": null\}\}, "required": ["query\_list"], "additionalProperties": null\}, "strict": false\}\}\par
\textless{}/tools\textgreater{}\par
\smallskip
For each function call, return a json object with function name and arguments within \textless{}tool\_call\textgreater{}\textless{}/tool\_call\textgreater{} XML tags:\par
\textless{}tool\_call\textgreater{}\par
\{"name": \textless{}function-name\textgreater{}, "arguments": \textless{}args-json-object\textgreater{}\}\par
\textless{}/tool\_call\textgreater{}\par
\end{quote}
\paragraph{User template}
\begin{quote}\small\ttfamily\raggedright
Answer the given question. You must conduct reasoning inside \textless{}think\textgreater{} and \textless{}/think\textgreater{} first every time you get new information. After reasoning, if you find you lack some knowledge, you can call a search engine by \textless{}tool\_call\textgreater{} query \textless{}/tool\_call\textgreater{} and it will return the top searched results between \textless{}tool\_response\textgreater{} and \textless{}/tool\_response\textgreater{}. You can search as many times as your want. If you find no further external knowledge needed, you can directly provide the answer inside \textless{}answer\textgreater{} and \textless{}/answer\textgreater{}, without detailed illustrations. For example, \textless{}answer\textgreater{} Beijing \textless{}/answer\textgreater{}. Question: \{question\}\par
\end{quote}

The tool is bound to the fixed KILT/E5 retriever, despite the generic web-search wording in its description. The native interface serialises calls as JSON objects with a \texttt{search} function and a \texttt{query\_list} argument; the user instruction's \texttt{<tool\_call> query </tool\_call>} is shorthand. The runtime enforces the four-assistant-turn limit independently of the prompt wording.

\paragraph{Tool interface and retrieval budget.}
Qwen3's native tool interface supports parallel tool calling. We retain this interface and execute only the first query in each search call to keep the retrieval budget per call fixed across conditions. This controls an additional source of variation while we compare reward construction and credit assignment. The tool response reports any ignored queries, which can be issued in later turns. All training conditions share this rule, and a trajectory may contain multiple calls.

\subsection{Reward and credit implementation}
\paragraph{Training conditions.}
OO uses terminal F1 with standard GRPO. Scalar variants of Cov, Cov-Dep, and AM use the same terminal F1 plus $0.25$ times the selected signal's accumulated new-maximum progress, followed by ordinary GRPO. Local variants retain terminal F1 as the outcome reward and add credit from one retrieval signal to the executed query tokens, with $\lambda=0.5$. Cov-local, Cov-Dep-local and AM-local apply this credit across all outcome groups, using the retrieval-progress events defined in Equation~\ref{eq:event}. Cov uses Equation~\ref{eq:unordered}; Cov-Dep adds the dependency closure in Equation~\ref{eq:graph}. AM-permute, AM-no-flat, and AM-flat-only intervene on AM-local. Cov-permute, Cov-no-flat, and Cov-flat-only intervene on Cov-local.

\paragraph{Normalisation and support.}
For eligible values $v_1,\ldots,v_m$, with $m\geq2$, define
\begin{equation}
 z(v)_i=\frac{v_i-\bar v}{\sqrt{\frac{1}{m-1}\sum_{j=1}^{m}(v_j-\bar v)^2}+\epsilon},
\end{equation}
where $\bar v=m^{-1}\sum_{j=1}^{m}v_j$ and $\epsilon=10^{-6}$. Outcome advantages use the complete rollout group for a question; local advantages use the eligible subset at the corresponding search step. If all centred values are within the numerical tolerance, the sample standard deviation is at most the tolerance, or the result is nonfinite, no local credit is assigned. We first standardise each signal and then standardise the combined local component. With one active signal, the second pass is numerically close to a single standardisation but still affects the result through the numerical tolerance. Comparisons use eligible searches at the same step across rollouts of the same question. Tokens must also be trainable under the response loss mask.

Execution records and token IDs associate each observation with its action and query span. A missing execution record, an unidentifiable query span, inconsistent event/span arrays, or a truncated trajectory disables local credit, leaving the global outcome advantage. Generated query text that is not executed still receives global supervision. Full payload length remains the mass reference, so event mass can depend on payload length.

\paragraph{Preserving credit under permutation.}
For an eligible event, $\sum_{k\in Q_{it}}\ell_{it}L_{it}/|Q_{it}|=\ell_{it}L_{it}$. The permutation variants transport the complete signed event mass before division by the recipient target length. This preserves the eligible actions and the multiset of signed event masses while changing which query receives each mass. The outcome-flat gate is computed from the complete group's terminal F1, not the subset that passes local eligibility. Removing a group's local term leaves its global term and sampled data unchanged.

\paragraph{Retrieval-event implementation.}
AM normalises titles, document bodies and nonempty reference answer aliases by lowercasing and removing punctuation, articles and redundant whitespace. It tests each document separately; matches are not assembled across documents. Query text, generated reasoning, final answers and tool-cap notices are excluded. Coverage potentials are rounded to four decimal places before conversion to new-maximum events.

\paragraph{Controlled comparisons.}
AM-permute and Cov-permute permute signed event masses among active, nonzero events for the same question at the same search step (seed 1729, fixed points allowed), retaining support and signed/absolute event mass. These controls break event--action correspondence. For both Cov and AM, no-flat retains local credit only when the complete group's outcome range exceeds $10^{-6}$; flat-only retains it only otherwise. These restrictions remove local advantage mass without rescaling the remaining updates. Section~\ref{sec:flat_groups} compares their results.

\subsection{Scalar reward scale}
\label{app:calibration}
We use $\alpha=0.25$ for all scalar conditions as a common trade-off between outcome reward and retrieval supervision. Both answer F1 and accumulated retrieval progress lie in $[0,1]$, so the retrieval bonus adds at most 0.25 to the combined reward. This allows retrieval progress to distinguish trajectories with similar answer scores while ensuring that an answer with F1 of one always outranks an answer with F1 of zero.

To check the effect on within-group reward rankings and advantage signs, we applied several coefficients to saved Cov-Dep-local and AM-local trajectories available before scalar training. The primary sample contains 768 five-rollout groups (3,840 trajectories) from short training runs; a supplementary sample contains 128 complete groups (640 trajectories) from later training stages. These fixed-trajectory comparisons informed the choice of reward scale without using evaluation scores.

\begin{table}[!htbp]
\caption{Effect of scalar reward coefficients on the primary sample. Rank reversal is measured over 2,617 within-group pairs with unequal original F1. Sign reversal is measured over 2,166 originally nonzero advantages in mixed-outcome groups. Both rates are computed by applying different coefficients to the same saved trajectories.}
\label{tab:calibration}
\centering\small
\begin{tabular}{rrrrr}
\toprule
& \multicolumn{2}{c}{Rank reversal (\%)} & \multicolumn{2}{c}{Advantage sign reversal (\%)} \\
$\alpha$ & Cov-Dep & AM & Cov-Dep & AM \\
\midrule
0.05 & 0.19 & 0.38 & 0.14 & 0.42 \\
0.10 & 0.38 & 0.76 & 0.42 & 0.92 \\
0.25 & 0.57 & 1.49 & 1.02 & 2.49 \\
0.50 & 0.73 & 2.45 & 2.03 & 4.39 \\
1.00 & 1.99 & 3.90 & 4.62 & 7.29 \\
\bottomrule
\end{tabular}
\end{table}

At $\alpha=0.25$, Cov-Dep and AM reverse 0.57\% and 1.49\% of unequal-F1 pairs, respectively, while their advantage sign-reversal rates are 1.02\% and 2.49\%. This setting preserves most outcome ordering in the saved local-policy sample. Mean accumulated retrieval progress in this sample is 0.2124 for Cov-Dep and 0.5393 for AM.

Scalar incorporation also supplies updates in outcome-flat groups. Let $I_i=\sum_t e_{it}$, with group mean $\bar I$ and sample standard deviation $s_I$. For a constant outcome $r_i=c$ and $\alpha>0$, the scalar advantage becomes
\begin{equation}
 z(c+\alpha I)_i=\frac{I_i-\bar I}{s_I+\epsilon/\alpha}.
\end{equation}
Except near the numerical tolerance, lowering $\alpha$ therefore does not proportionally lower this standardised update. Of 332 outcome-flat groups in the primary sample, Cov-Dep activates 63 and AM activates 49 for every inspected positive coefficient. Among 183 all-zero-F1 groups, each signal activates 35 groups. Among the 77 tied-outcome groups in the supplementary sample, Cov-Dep produces nonzero advantages in 14 groups and AM in 5. The equation and observed activations show that scalar bonuses also provide a learning signal in outcome-flat groups.

\subsection{Evaluation and trajectory measurements}
\label{app:dev_coverage}
\paragraph{Evaluation splits and aggregation.}
The main evaluation contains 3,197 questions: six datasets with 512 questions each and Bamboogle with 125. Dataset-level EM and F1 scores are weighted by their sample counts, so each evaluation question contributes equally to Avg.

\paragraph{Development trajectories.}
We evaluate OO and Cov-local on the same 1,200 synthetic development questions. For each question, we retain the sequence of assistant messages, search actions, tool observations and the final answer. We map retrieved text in complete visible tool observations to the original corpus chunks and apply the coverage predicate from Section~\ref{sec:construction}. Generated reasoning and answers do not contribute to coverage. For this analysis, we use the exact fraction of covered gold nodes rather than the rounded values used to construct training rewards. Figure~\ref{fig:dev_coverage} counts the questions at each covered-node proportion. We compute mean coverage and search count over all 1,200 questions. Figure~\ref{fig:dev_coverage_f1} groups each policy's questions separately by final Cov and averages answer F1 within each group; it does not filter for matched question IDs at a given coverage value.

All questions remain in the denominator, including missing answers and truncated trajectories. OO has 19 missing answers and 12 truncated trajectories; Cov-local has 20 and 16, respectively. One OO trajectory ends with an incomplete tool observation, which is excluded from coverage; executed searches still contribute to the search count. Search count measures tool use, rather than generation cost or wall-clock latency. As detailed in the data-construction appendix, 507 development questions share some annotated evidence with training, while 693 have fully unseen annotated evidence. This analysis examines behaviour on the development distribution separately from the seven-benchmark results.

\section{Training data and corpus construction}
\label{app:data}
\paragraph{Construction constraints.}
We use the August 2019 KILT Wikipedia corpus and the August 2022 Wikidata truthy dump. For chain composition, we retain properties with at most 1.5 objects per subject on average, and require the selected subject--property pair to have exactly one object. For a conjunction, both branches must have multiple candidate objects and their intersection must contain exactly one. These constraints define the annotated graph structure. Relations are selected from corpus co-occurrences and filtered using Wikidata; each selected relation retains its supporting Wikipedia passage.

\paragraph{Generation model.}
We use GPT-5.6 Luna (\texttt{gpt-5.6-luna}) for root-description generation throughout the source pool. Graph operations and templates determine the question structure, answers and dependencies.

\paragraph{Evidence units and chunk mapping.}
Several atomic nodes can share a supporting paragraph. We merge shared supporting paragraphs when assigning structural categories for sampling, while preserving the original atomic nodes and prerequisites used by Cov-Dep and Cov. The resulting categories describe annotated evidence structure rather than the agent's search count. To map evidence to indexed chunks, we remove the page's initial title entry, retain section and bullet marker lines, join the remaining text with newlines, and split it into non-overlapping 100-word chunks. Grounding normalises page and paragraph identifiers and locates answer mentions within these spans; the coverage rule in Section~\ref{sec:construction} also checks page identity.

\paragraph{Mixture selection.}
We sample deterministically within each structural category, reserving 1,200 questions for the evidence-coverage analysis in Section~\ref{sec:dev_coverage} and selecting 14,000 training questions from the remainder. We refer to the reserved questions as the development split. Of 87,171 source questions, 86,681 are eligible: 168 fail the chunk-grounding eligibility check, 297 are labelled graph shortcuts by the construction filter, and 25 do not match a requested category. For the first four categories, half the selected records have all gold evidence in the first chunk of each supporting page and half require later chunks. Other categories are sampled from their full eligible pools.

\begin{table}[!htbp]
\caption{Training and development counts by construction category. Structural labels follow the naming convention of MuSiQue \citep{trivedi-etal-2022-musique}; numbered suffixes distinguish graph templates at the same hop count. Categories account for shared supporting paragraphs and describe evidence structure, not search counts.}
\centering\small
\begin{tabular}{lrr}
\toprule
Construction category & Train & Dev \\
\midrule
1-hop, single & 1,200 & 150 \\
1-hop, conjunctive & 1,300 & 150 \\
2-hop, chain & 2,400 & 150 \\
2-hop, wide & 900 & 100 \\
3-hop-1, chain & 2,000 & 150 \\
3-hop-1, wide & 700 & 50 \\
3-hop-2, fan-in & 2,200 & 150 \\
4-hop-1, chain & 1,000 & 100 \\
4-hop-2, fan-in & 1,300 & 100 \\
4-hop-3, fan-in & 1,000 & 100 \\
\midrule
Total & 14,000 & 1,200 \\
\bottomrule
\end{tabular}
\end{table}

\paragraph{Corpus grounding.}
Each record contains a question, answer aliases, a decomposition, dependency references, and corpus grounding. The selected data contain 44,385 graph nodes, including 8,050 conjunctive nodes. The retrieval index contains approximately 29 million chunks represented by normalised 768-dimensional embeddings. Every node is grounded in the corpus, every prerequisite refers to an existing node, and the dependency graphs are acyclic.

\paragraph{Matching annotated evidence to retrieved chunks.}
Each node stores a list of required source passages, identified by Wikipedia page and paragraph, and the character spans of its subquestion-answer mentions in each passage. For every required passage, the detector checks whether any retrieved chunk has the same page and paragraph and contains the start offset of at least one annotated mention. A node is covered only if this test succeeds for all of its required passages. Different passages may be matched by different chunks and at different search steps. Both Cov and Cov-Dep use this check; dependency closure is applied separately by Cov-Dep. The check does not require the entire answer mention or supporting sentence to fit inside the chunk, as quantified below.

\paragraph{Question templates and split overlap.}
All 15,200 selected questions begin with the same ``What is the name of'' surface template. Normalised answer-alias strings occur in 57 training and 5 development questions. These matches include ordinary-word collisions and aliases embedded in longer names, as well as potential answer cues. There is no exact question-ID or normalised-question overlap between the two splits, but 225 development questions have all gold evidence seen in training and 282 have partial evidence overlap; only 693 have fully unseen gold evidence. The development split contains 531 questions sharing a training page and 606 sharing an answer entity. This split is reserved separately from the benchmark evaluation; the shared evidence limits its value for measuring generalisation.

\paragraph{Partial mention coverage.}
All 52,435 gold annotation records have start-position coverage. For 265 records, no single chunk fully contains any of the annotated answer mentions. Across matched annotation--chunk pairs, 271 contain a mention's start but no complete annotated mention. Chunk identification uses a text prefix, so identifying a chunk does not establish that its full text is present in the observation.

\paragraph{Structural and grounding checks.}
We verify graph validity and the alignment between evidence annotations and retriever chunks. Table~\ref{tab:instrumentation} summarises these properties and the effects of chunk boundaries and dependency constraints on retrieval progress.

\begin{table}[!htbp]
\caption{Graph structure and chunk grounding in the training and development data, with retrieval-progress differences measured on 7,680 training trajectories.}
\label{tab:instrumentation}
\centering\small
\begin{tabularx}{\linewidth}{Yr}
\toprule Check & Observed count \\
\midrule
Selected train + development records & 15,200 \\
Grounded graph nodes (conjunctive subset) & 44,385 (8,050) \\
Missing prerequisites / cycles / missing node grounding & 0 / 0 / 0 \\
Gold annotation records with start-position coverage & 52,435 / 52,435 \\
Gold annotation records without full-mention chunk coverage & 265 / 52,435 \\
Trajectories affected by requiring full-mention coverage & 20 / 7,680 \\
Trajectories with Cov progress above Cov-Dep at some point & 310 / 7,680 \\
Trajectories exhibiting delayed unlocking & 21 / 7,680 \\
\bottomrule
\end{tabularx}
\end{table}
The trajectory measurements use 7,680 training trajectories collected before the final-policy comparisons.

\paragraph{Effect of chunk boundaries on retrieval progress.}
Table~\ref{tab:instrumentation} shows how chunk boundaries and dependency constraints affect retrieval progress. Requiring the complete answer mention to fit within a retrieved chunk reduces Cov-Dep progress on 20 of the 7,680 trajectories. Dependency constraints affect which matched nodes receive credit: Cov exceeds Cov-Dep on 310 trajectories, while 21 contain evidence credited only after a later search retrieves its prerequisites.

Grounding determines whether a retrieved chunk matches an annotated node; closure determines when that match contributes to progress. The two checks affect the same reward at different stages.

\section{Additional results}
\label{app:results}
Tables~\ref{tab:fullf1} and~\ref{tab:em} report F1 and exact-match accuracy for all 13 conditions, including the local-credit interventions summarised in Figure~\ref{fig:mechanisms}. All results follow the evaluation protocol in Section~\ref{sec:experiments}.

\begin{table}[!htbp]
\caption{Complete answer F1 (\%) results for all 13 conditions on seven datasets. Avg follows the aggregation used in Table~\ref{tab:main}; bold marks the highest value in each column across all conditions.}
\label{tab:fullf1}
\centering\small\setlength{\tabcolsep}{3pt}
\begin{tabular}{lrrrrrrrr}
\toprule
Method & 2Wiki & HotpotQA & MuSiQue & Bamboogle & NQ & TriviaQA & PopQA & Avg \\
\midrule
OO & 57.30 & 53.74 & 24.93 & 57.73 & 42.04 & 74.46 & 53.45 & 51.25 \\
\midrule
Cov-scalar & 59.26 & 57.54 & 25.69 & 59.51 & 43.08 & 74.09 & 54.50 & 52.64 \\
Cov-Dep-scalar & 57.81 & 58.42 & 25.11 & 59.54 & \textbf{44.51} & 74.58 & \textbf{54.88} & 52.82 \\
AM-scalar & 59.27 & 57.11 & 25.89 & 60.21 & 43.25 & 74.45 & 54.16 & 52.66 \\
\midrule
Cov-local & \textbf{62.82} & 58.91 & \textbf{28.84} & \textbf{61.91} & 43.43 & 75.86 & 54.37 & \textbf{54.34} \\
Cov-Dep-local & 60.96 & \textbf{59.49} & 25.49 & 58.29 & 42.01 & 76.09 & 52.42 & 52.96 \\
AM-local & 59.26 & 59.13 & 27.23 & 59.01 & 43.23 & \textbf{76.31} & 54.57 & 53.51 \\
\midrule
AM-permute & 52.99 & 53.83 & 24.13 & 55.99 & 42.45 & 75.63 & 51.95 & 50.39 \\
Cov-permute & 57.36 & 57.10 & 25.22 & 56.90 & 43.33 & 75.02 & 54.14 & 52.22 \\
\midrule
Cov-no-flat & 61.27 & 58.85 & 28.77 & 60.06 & 43.57 & 75.51 & 54.69 & 54.02 \\
Cov-flat-only & 59.56 & 54.61 & 25.32 & 57.82 & 42.67 & 75.33 & 53.53 & 52.07 \\
AM-no-flat & 56.05 & 57.87 & 26.72 & 53.57 & 42.46 & 75.15 & 53.82 & 52.07 \\
AM-flat-only & 57.60 & 57.37 & 23.00 & 58.26 & 42.52 & 74.45 & 51.26 & 51.32 \\
\bottomrule
\end{tabular}
\end{table}

\begin{table}[!htbp]
\caption{Exact match (\%) on all seven datasets. Avg follows the aggregation used in Table~\ref{tab:main}; bold marks the highest value in each column.}
\label{tab:em}
\centering\small\setlength{\tabcolsep}{3pt}
\begin{tabular}{lrrrrrrrr}
\toprule
Method & 2Wiki & HotpotQA & MuSiQue & Bamboogle & NQ & TriviaQA & PopQA & Avg \\
\midrule
OO & 45.90 & 41.80 & 13.28 & 46.40 & 24.22 & 62.89 & 45.51 & 39.22 \\
\midrule
Cov-scalar & 48.44 & 45.51 & 15.82 & 44.80 & 26.76 & 63.09 & 46.68 & 41.19 \\
Cov-Dep-scalar & 46.48 & 46.48 & 15.04 & 46.40 & \textbf{27.73} & 64.45 & \textbf{47.85} & 41.54 \\
AM-scalar & 47.66 & 45.12 & 14.65 & 48.00 & \textbf{27.73} & 63.87 & 47.07 & 41.29 \\
\midrule
Cov-local & \textbf{51.95} & 46.88 & 16.60 & \textbf{49.60} & 26.56 & \textbf{65.43} & 46.68 & \textbf{42.63} \\
Cov-Dep-local & 49.80 & 46.88 & 15.43 & 48.00 & 25.00 & 65.04 & 44.73 & 41.41 \\
AM-local & 49.02 & \textbf{47.07} & 16.21 & 46.40 & 25.78 & 64.84 & 46.88 & 41.82 \\
\midrule
AM-permute & 41.99 & 41.41 & 13.09 & 43.20 & 25.98 & 65.04 & 44.14 & 38.79 \\
Cov-permute & 47.27 & 45.70 & 14.84 & 45.60 & 27.54 & 64.26 & 46.68 & 41.23 \\
\midrule
Cov-no-flat & 51.17 & 46.68 & \textbf{17.38} & 46.40 & 26.56 & 65.04 & 46.88 & 42.45 \\
Cov-flat-only & 48.44 & 42.97 & 13.48 & 44.80 & 25.00 & 64.26 & 46.09 & 40.23 \\
AM-no-flat & 45.51 & 45.70 & 16.60 & 43.20 & 25.98 & 64.45 & 46.88 & 40.94 \\
AM-flat-only & 46.48 & 44.73 & 12.70 & 46.40 & 25.39 & 63.67 & 43.55 & 39.69 \\
\bottomrule
\end{tabular}
\end{table}

\begin{figure}[t]
\centering\includegraphics[width=\linewidth]{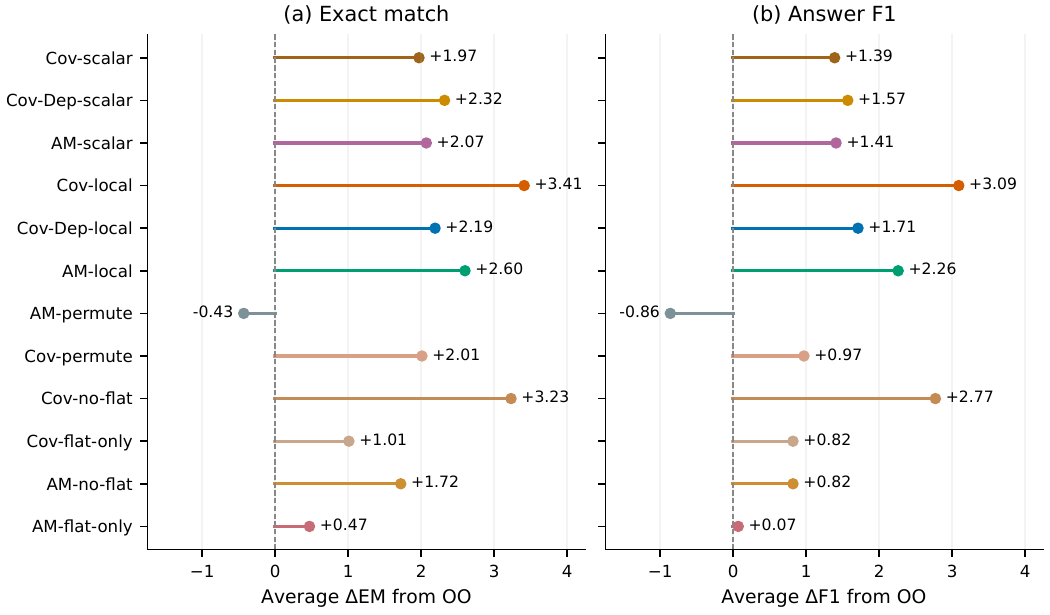}
\caption{Differences in average EM and F1 from outcome-only training, on the same score scale as Table~\ref{tab:main}. Both scalar and local incorporation improve EM and F1 over OO for all three signals; Cov-local gives the largest gains.}
\label{fig:finalcomparison}
\end{figure}

\begin{figure}[t]
\centering\includegraphics[width=\linewidth]{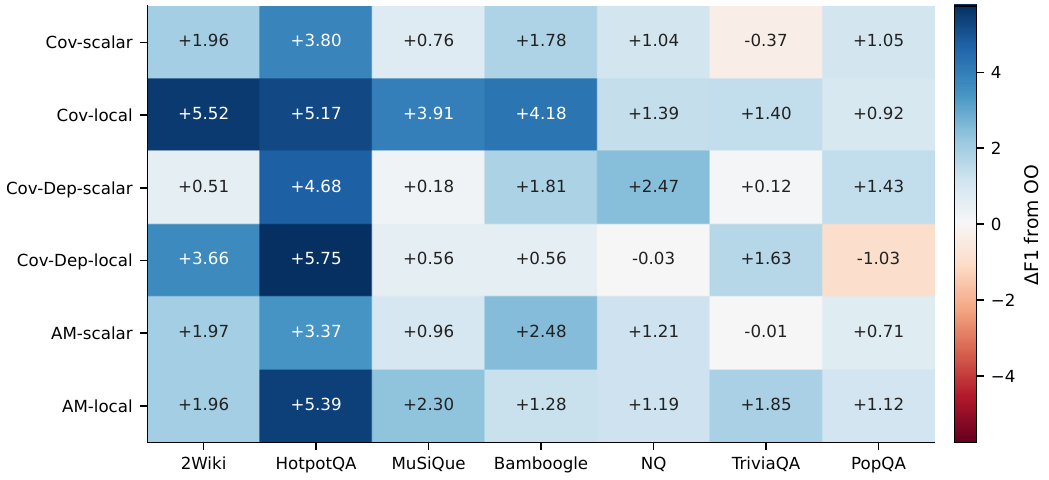}
\caption{Per-dataset F1 differences from OO, on the same score scale as Table~\ref{tab:main}. The heatmap shows how each retrieval signal and incorporation rule affects individual benchmarks.}
\label{fig:datasetdeltas}
\end{figure}

\subsection{Multi-hop and single-hop QA}
\label{app:taskgroups}
We group 2WikiMultiHopQA, HotpotQA, MuSiQue, and Bamboogle as multi-hop QA (1,661 examples), and Natural Questions, TriviaQA, and PopQA as single-hop QA (1,536 examples). Table~\ref{tab:taskgroups} averages per-question scores within each group, retaining the original dataset sizes as weights. Bamboogle contributes 125 examples, and each other dataset contributes 512.

\begin{table}[!htbp]
\caption{F1 (\%) within each QA task group and across all 3,197 examples. Each question has equal weight within the reported average. Bold marks the highest value in each column.}
\label{tab:taskgroups}
\centering\small
\begin{tabular}{lrrr}
\toprule Method & Multi-hop & Single-hop & All \\
\midrule
OO & 46.26 & 56.65 & 51.25 \\
\midrule
Cov-scalar & 48.40 & 57.22 & 52.64 \\
Cov-Dep-scalar & 48.05 & 57.99 & 52.82 \\
AM-scalar & 48.38 & 57.28 & 52.66 \\
\midrule
Cov-local & \textbf{51.07} & 57.88 & \textbf{54.34} \\
Cov-Dep-local & 49.37 & 56.84 & 52.96 \\
AM-local & 49.33 & \textbf{58.03} & 53.51 \\
\midrule
AM-permute & 44.58 & 56.68 & 50.39 \\
Cov-permute & 47.34 & 57.49 & 52.22 \\
\midrule
Cov-no-flat & 50.41 & 57.93 & 54.02 \\
Cov-flat-only & 47.35 & 57.18 & 52.07 \\
AM-no-flat & 47.38 & 57.14 & 52.07 \\
AM-flat-only & 46.91 & 56.07 & 51.32 \\
\bottomrule
\end{tabular}
\end{table}

The query-alignment contrasts are larger on multi-hop QA: Cov-local exceeds Cov-permute by 3.73 F1 on multi-hop QA and 0.39 F1 on single-hop QA; AM-local exceeds AM-permute by 4.75 and 1.35 F1, respectively. Correct correspondence therefore improves both task groups under both signals, with larger differences on multi-hop QA. All three scalar variants improve over OO in both task groups. Local incorporation further improves multi-hop QA for every signal, while on single-hop QA Cov-Dep-local falls below its scalar counterpart. The groups contain different benchmarks, so this comparison does not isolate the effect of question hop count.

\subsection{Sensitivity to the dataset mixture}
\label{app:sensitivity}
Table~\ref{tab:leaveone} recomputes two F1 contrasts after omitting each dataset in turn. We retain the original sample counts as weights for the remaining six datasets. Cov-local and AM-local remain above OO under every omission. This check measures sensitivity to the dataset mixture.

\begin{table}[!htbp]
\caption{Changes in average F1 after omitting one dataset. The full seven-dataset contrasts are $+3.09$ and $+2.26$, respectively.}
\label{tab:leaveone}
\centering\small
\begin{tabular}{lrr}
\toprule Omitted dataset & Cov-local $-$ OO & AM-local $-$ OO \\
\midrule
2Wiki & +2.63 & +2.32 \\
HotpotQA & +2.70 & +1.66 \\
MuSiQue & +2.94 & +2.26 \\
Bamboogle & +3.05 & +2.30 \\
NQ & +3.42 & +2.47 \\
TriviaQA & +3.42 & +2.34 \\
PopQA & +3.51 & +2.48 \\
\bottomrule
\end{tabular}
\end{table}

\section{Coverage example}
\label{app:coverage_example}

\paragraph{Retrieved evidence in Figure~\ref{fig:pipeline}a.}
The question asks which dynasty the father of the Burmese monarch who ruled from 1171 to 1174 belonged to. The displayed rollout searches for \emph{Burma monarch 1171-1174 father king dynasty}. Its first search returns passages titled Narathu, Naratheinkha and Saw Lu. The Naratheinkha passage matches the first annotated node, identifying the monarch from his reign; the Narathu passage matches the third node, identifying the dynasty. The second node requires a different passage establishing Naratheinkha's father and remains unmatched. Thus the covered set is $\{v_1,v_3\}$, while dependency closure admits only $\{v_1\}$. The first-search potentials and events are both 0.6667 for Cov and 0.3333 for Cov-Dep, displayed as 0.67 and 0.33 in the figure. The matched dynasty fact is credited by Cov even though the intervening father link has not been retrieved.

This example illustrates the detector comparison; Figure~\ref{fig:pipeline}b shows the general credit rules schematically. We examine the saved trajectories in a fixed order and select the first whose initial search satisfies $0<\mathrm{Cov\text{-}Dep}<\mathrm{Cov}$, without filtering on answer correctness.

\section{A paired example of evidence acquisition}
\label{app:trajectory_case}
Figure~\ref{fig:trajectory_case} compares the two policies on a question about Robert A. Rosenstone's institution. OO's first two searches return other scholars; its third retrieves an introductory passage that identifies Rosenstone and names Caltech. Cov-local retrieves that passage on its first search, then obtains an additional annotated career passage on its second. Both final answers are correct. The additional passage raises annotated coverage from one node to two, even though the introductory passage already contains the answer. This case illustrates the distinction between recovering an answer and covering the annotated evidence.

We select the first question in lexicographic ID order among complete multi-node pairs where Cov-local attains full annotated coverage, OO does not, and Cov-local executes fewer searches. Three pairs satisfy these criteria; correctness and F1 are not selection criteria. The figure retains the executed queries and final response text, with observation descriptions abridged against the original returned passages.

\begin{figure}[!htbp]
\centering
\begin{minipage}{\linewidth}
\small
\textbf{Question.} What is the name of the institution where an American historian who studies the relationship between history and visual media has been Professor of History since 1966?\par\medskip
\textbf{Annotated evidence.} Identify Robert A. Rosenstone; retrieve the annotated career passage naming his institution.\par\medskip
\begin{minipage}[t]{0.47\linewidth}
\textbf{OO}\par\smallskip\hrule\smallskip
\textbf{Search 1.} \emph{American historian studies history and visual media Professor since 1966}\par
Returns passages about other historians and scholars; neither annotated node is covered. \textbf{Cov 0.00.}\par\medskip
\textbf{Search 2.} \emph{American historian history and visual media Professor since 1966 institution}\par
Again returns other scholars; neither annotated node is covered. \textbf{Cov 0.00.}\par\medskip
\textbf{Search 3.} \emph{historian who studies history and visual media Professor since 1966 institution}\par
Identifies Robert A. Rosenstone and names Caltech, covering the first node. \textbf{Cov 0.50.}\par\medskip
\textbf{Final response (verbatim).}\par
The American historian who studies the relationship between history and visual media and has been a Professor of History since 1966 is Robert A. Rosenstone. He is a leading scholar in the field of the relationship between history and visual media, having authored books such as *Visions of the Past: The Challenge of Film to Our Idea of History* and *History on Film / Film on History*. He is Professor Emeritus of History at the California Institute of Technology (Caltech).

Thus, the institution is the **California Institute of Technology**.\par\smallskip
\textbf{Answer: California Institute of Technology.}\par
\end{minipage}
\hfill
\begin{minipage}[t]{0.47\linewidth}
\textbf{Cov-local}\par\smallskip\hrule\smallskip
\textbf{Search 1.} \emph{American historian who studies the relationship between history and visual media has been Professor of History since 1966}\par
Identifies Robert A. Rosenstone and names Caltech, covering the first node. \textbf{Cov 0.50.}\par\medskip
\textbf{Search 2.} \emph{Robert A. Rosenstone Professor Emeritus of history at California Institute of Technology since 1966}\par
Also retrieves the annotated career passage: “He was a Professor of History at the Caltech”. Both nodes are covered. \textbf{Cov 1.00.}\par\medskip
\textbf{Final response (verbatim).}\par
The American historian who studies the relationship between history and visual media and has been a Professor of History since 1966 is Robert A. Rosenstone. He became a Professor of History at the California Institute of Technology (Caltech) in 1966 and is described as the leading international scholar in the field of history and visual media.\par\smallskip
\textbf{Answer: California Institute of Technology.}\par
\end{minipage}
\end{minipage}
\caption{Two search trajectories for the same development question. Queries are the executed queries verbatim; observation descriptions are abridged, with the direct evidence quoted. Both policies answer correctly, but their retrieved support and search counts differ.}
\label{fig:trajectory_case}
\end{figure}

\end{document}